\documentclass[letterpaper]{article} 
\usepackage{aaai24}  
\usepackage{times}  
\usepackage{helvet}  
\usepackage{courier}  
\usepackage[hyphens]{url}  
\usepackage{graphicx} 
\usepackage{natbib}  
\usepackage{caption} 
\usepackage{algorithm}
\usepackage{algorithmic}
\usepackage{multirow}
\usepackage{newfloat}
\usepackage{listings}
\DeclareCaptionStyle{ruled}{labelfont=normalfont,labelsep=colon,strut=off} 
\floatstyle{ruled}
\newfloat{listing}{tb}{lst}{}
\floatname{listing}{Listing}
\title{AI-Based Energy Transportation Safety: \\ Pipeline Radial Threat Estimation Using Intelligent Sensing System}
\author{
    Chengyuan Zhu\textsuperscript{\rm 1},
    Yiyuan Yang\textsuperscript{\rm 2}\thanks{Work done as a student in Tsinghua University. He is now with the Department of Computer Science, University of Oxford, OX1 3SA, Oxford, UK.}, 
    Kaixiang Yang\textsuperscript{\rm 3}\footnote{Corresponding author},
    Haifeng Zhang\textsuperscript{\rm 4},\\
    Qinmin Yang\textsuperscript{\rm 1}\footnotemark[2],
    C. L. Philip Chen\textsuperscript{\rm 3}
}
\affiliations{
    \textsuperscript{\rm 1}Zhejiang University, Hangzhou, China, \textsuperscript{\rm 2}University of 
Oxford, Oxfordshire, United Kingdom, \textsuperscript{\rm 3}South China University of Technology, Guangzhou, China, \textsuperscript{\rm 4}Research Institute of Tsinghua University, Pearl 
River Delta, Guangzhou, China

    zhuchengyuan517@zju.edu.cn, yiyuan.yang@cs.ox.ac.uk, yangkaixiang@zju.edu.cn,\\
    zhanghf@tsinghua-gd.org,
    qmyang@zju.edu.cn,
    philipchen@scut.edu.cn
}

\usepackage{bibentry}

\begin{document}

\maketitle

\begin{abstract}
The application of artificial intelligence technology has greatly enhanced and fortified the safety of energy pipelines, particularly in safeguarding against external threats. The predominant methods involve the integration of intelligent sensors to detect external vibration, enabling the identification of event types and locations, thereby replacing manual detection methods. However, practical implementation has exposed a limitation in current methods - their constrained ability to accurately discern the spatial dimensions of external signals, which complicates the authentication of threat events. Our research endeavors to overcome the above issues by harnessing deep learning techniques to achieve a more fine-grained recognition and localization process. This refinement is crucial in effectively identifying genuine threats to pipelines, thus enhancing the safety of energy transportation. This paper proposes a radial threat estimation method for energy pipelines based on distributed optical fiber sensing technology. Specifically, we introduce a continuous multi-view and multi-domain feature fusion methodology to extract comprehensive signal features and construct a threat estimation and recognition network. The utilization of collected acoustic signal data is optimized, and the underlying principle is elucidated. Moreover, we incorporate the concept of transfer learning through a pre-trained model, enhancing both recognition accuracy and training efficiency. Empirical evidence gathered from real-world scenarios underscores the efficacy of our method, notably in its substantial reduction of false alarms and remarkable gains in recognition accuracy. More generally, our method exhibits versatility and can be extrapolated to a broader spectrum of recognition tasks and scenarios.

\end{abstract}

\section{Introduction}

Energy safety has a significant impact on natural safety and social stability. Among them, energy transportation is an important part of energy safety. Pipelines undertake the majority of energy transportation capacity, including oil, natural gas, hydrogen, etc \cite{r1}. Especially in the oil and gas field, according to statistics, 70\% of crude oil in the United States is transported through pipelines, and the total length of global energy pipelines has exceeded 4 million kilometers. However, long-distance energy pipelines are in open natural areas and pose great safety risks. Pipeline leakage and breakage accidents may threaten human life, property and health, and have a negative and nonnegligible social impact. In 2022, the Nord Stream natural gas pipeline suffered external damage and leaked, resulting in multiple explosions\footnote{https://en.wikipedia.org/wiki/2022\_Nord\_Stream\_pipeline\_sabotage}. In the same year, more than 14,000 barrels of crude oil leaked from the pipeline of Kansas, USA, causing significant economic loss and environmental pollution\footnote{https://en.wikipedia.org/wiki/2022\_Keystone\_Pipeline\_oil\_spill}. 

To ensure pipeline safety, a large amount of manpower and resources have been invested in pipeline inspection recently. The development of intelligent sensing, data-driven, and advanced artificial intelligence (AI) technology has empowered the pipeline safety field \cite{r2, rykx1, ryqm1}. Therefore, industry experts and researchers have begun to combine sensing measurement technology with AI algorithms to achieve pipeline safety early warning (PSEW) \cite{rzhu3, ricann23, ryangtim}. Among them, distributed optical fiber sensing (DOFS) technology plays a crucial role and has been widely recognized by the industry. On the one hand, it has the advantages of high sensitivity, long-distance continuous monitoring, low cost, anti-electromagnetic interference, etc. On the other hand, the device is easy to deploy in stations along the pipeline. The storage and invocation of data, as well as the training and updating of algorithms, have become convenient \cite{rzhureview, rcac22}.

\begin{figure*}[!t]
\centering
\includegraphics[width=0.943\textwidth]{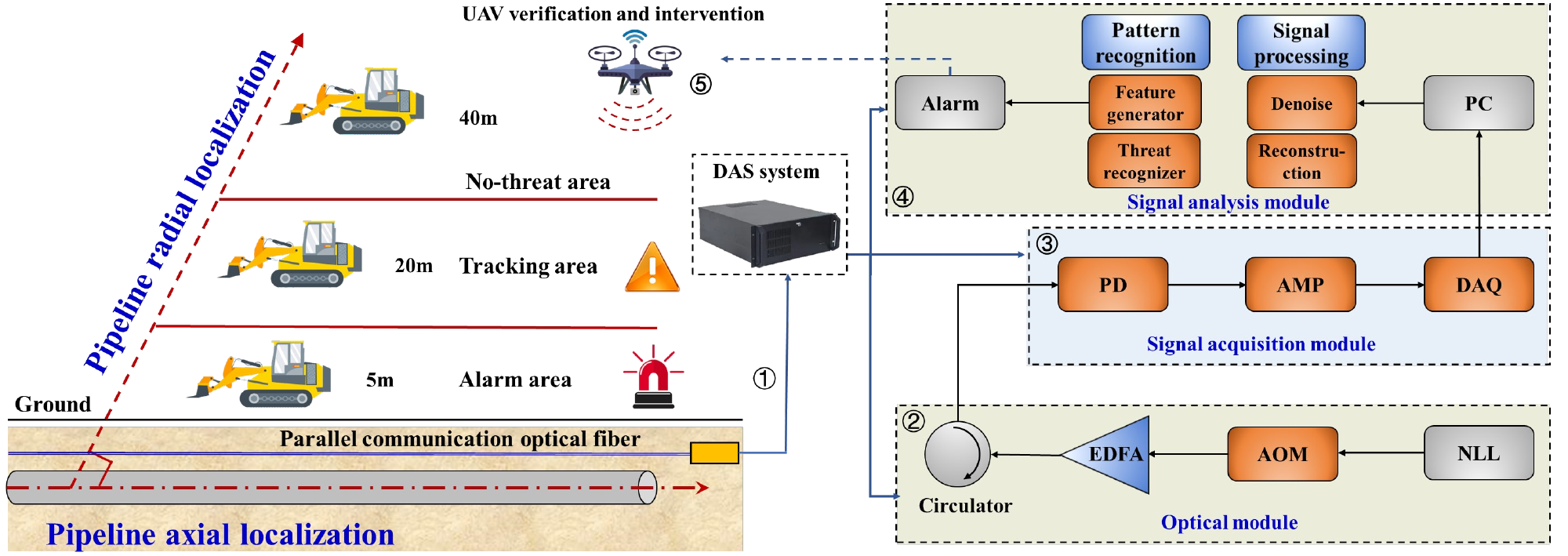}
\caption{The structure of an typical optical-fiber-based PSEW facility and architecture of the DAS system.}
\label{fig_1}
\end{figure*}

However, the current development of this technology still faces many important challenges. (1) There are complex background noise and interference signals in the external environment of the pipeline, making it difficult to distinguish similar signals. 
(2) The identification and location of threat events are only based on the pipeline axial location (i.e., defense zone), and it is difficult to determine the radial location of the threat event from the pipeline, which places higher requirements on spatial recognition ability. 
(3) The false alarm rate (FAR) is quite high, with a large number of false early warnings occurring in the radial direction of the pipeline, which wastes a lot of verification costs and is not conducive to optimizing scheduling and other maintenance measures.

These challenges greatly limit the accuracy and efficiency of DOFS technology in safety monitoring and warning for energy pipelines. To address the challenges currently faced by DOFS technology, this paper uses an advanced deep learning algorithm framework to build a fine-grained pipeline radial threat recognition algorithm, which helps pipeline workers more accurately judge and process warning results. Specifically, we first collect data from real sites and preprocess the original signal of fiber optic sensing to enhance signal quality, including signal denoising and reconstruction. Secondly, we integrate multi-view and multi-domain features from both the time-domain, frequency-domain and space-domain of continuous defense zones in feature construction. Next, the EfficientNetv2 model with the introduction of transfer learning is used to classify the threat level and identify the external damage events in the real danger area. Finally, the processing decisions of external threats to pipelines are subdivided with experts in the energy pipeline field, and unmanned aerial vehicle (UAV) verification and intervention are taken to the real threats.

Our main contributions are summarized as follows.

\begin{itemize}
    \item Compared with previous work, we focus on fine-grained identification of PSEW issues. We enhance the early warning system by estimating the distance of external damage events along the pipeline's radial direction, effectively mining valuable features from sensing signals.
    \item We innovate a preprocessing technique for extracting multi-view, multi-domain features from continuous space-time-frequency data. This approach maximizes the potential of optical fiber acoustic signals gathered during real-world scenarios. Our experiments confirm the superiority of our method over alternatives.
    \item We elevate model performance and classification outcomes through data augmentation, composite scale feature extraction, and pre-training models. This also accelerates model training, expediting convergence.
    \item Our method accurately recognizes safety threats from external pipeline damage, minimizing operational expenses while ensuring timely alerts. This contributes to a stable energy supply and societal progress.
\end{itemize}

\section{Related Works}

\subsection{Distributed Optical Fiber Sensing for PSEW}

\begin{figure*}[!t]
\centering
\includegraphics[width=0.943\textwidth]{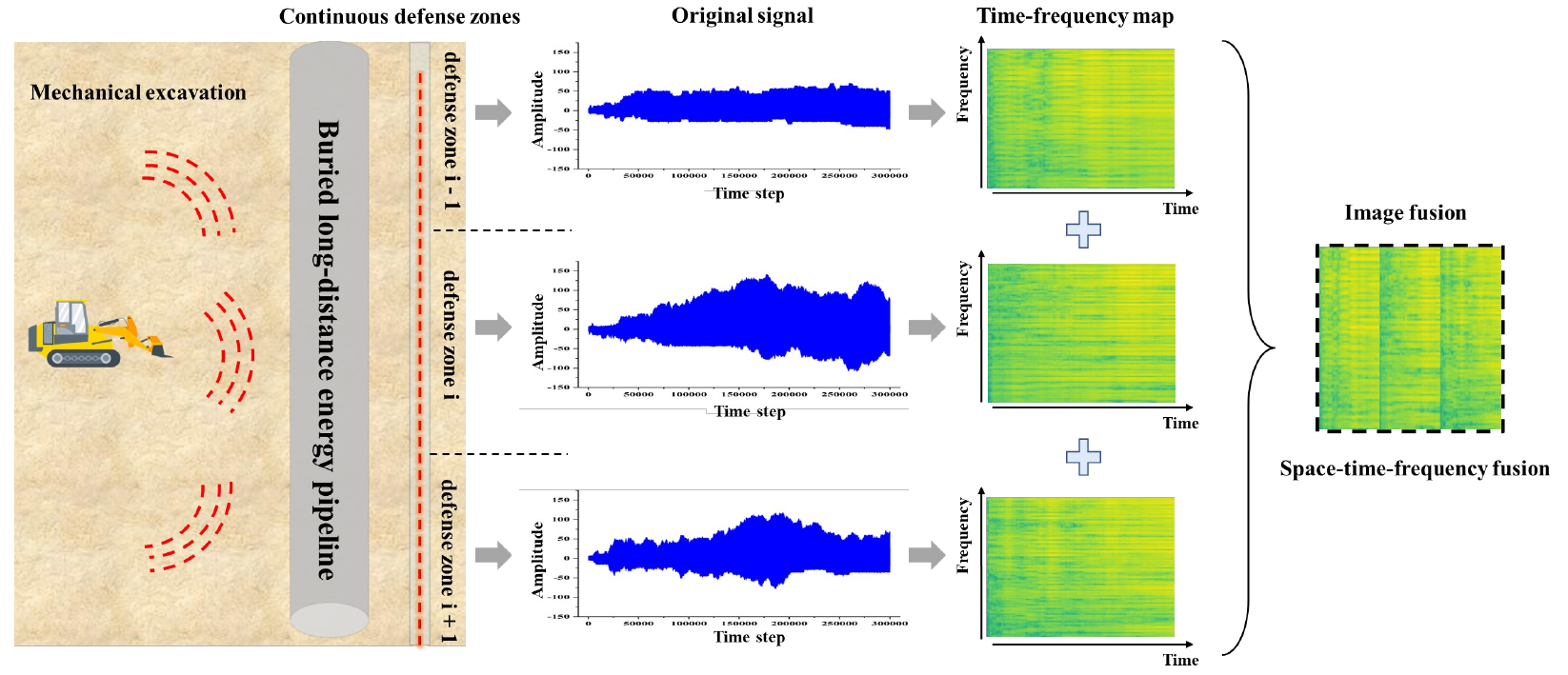}
\caption{The feature construction of multi-view and multi-domain based on space-time-frequency information fusion.}
\label{fig_2}
\end{figure*}

DOFS has the advantages of high sensitivity, distributed detection, and high signal-to-noise ratio of vibration signals \cite{ryangicassp}. For long-distance monitoring, the measurement range and spatial resolution of the system are two important indicators that need to be balanced. Moreover, the DOFS system used for pipelines needs a longer measurement range and usually needs to continuously monitor tens to hundreds of kilometers, so the spatial resolution is low. In the PSEW field \cite{rpd1, rpms1}, it is hoped that the DOFS system has both a high accuracy rate and a low false alarm rate to reduce the frequency of false alarms, so as to reduce the operation and maintenance cost of the measures taken after each early warning. However, due to the complex environment of long-distance pipeline \cite{rzhuswint, rlzl}, it is still a challenging problem in practical applications to classify and identify various vibrations with a high recognition rate.

Distributed acoustic sensing (DAS) \cite{r35} is a novel technology that enables continuously distributed detection of vibrations and acoustic fields. In the intrusion detection methods of the DAS system, the pattern recognition algorithms can classify vibration events \cite{rpr1, rpr2}. Yang et al. constructed a multi-feature fusion convolution neural network (CNN) and LightGBM model based on two new complementary spatial-temporal features and achieved an effective recognition rate of more than 95\% in the real environment \cite{ryangtim}. More effective feature extraction was the key to improving the recognition accuracy \cite{rfe1, rykx2}. New research began to try feature extraction and recognition technology based on multidimensional signals \cite{rpr3, rfe2, raaai1}. Meanwhile, the real-time transformation of massive raw data into useful sensing signals is an important direction for future research \cite{ryangtim, rzk}. In general, DOFS provides a transformative means for the perception of the physical world and has significant implications for advancement.

\subsection{Spatial Localization for Pipeline External Threats}

As shown in Fig. \ref{fig_1}, spatial localization for pipeline external threats can be divided into two main parts, pipeline axial and radial localization. The axial direction corresponds to the defense zone in the DOFS system. The radial direction represents the direction perpendicular to the pipeline. Fig. \ref{fig_1} demonstrates the intelligent identification, localization and decision-making intervention process of the entire PSEW system. First, the optical module (\textcircled{2}) of the DAS system obtains the acoustic signal along the external pipeline through the parallel communication optical fiber (\textcircled{1}). The parallel communication optical fiber is laid along with the pipeline construction. The optical module is responsible for transmitting optical signals. Second, the DAS system obtains and analyzes the perception signal in real time through the signal acquisition module (\textcircled{3}) and the signal analysis module (\textcircled{4}). Among them, the signal acquisition module realizes the conversion and digital demodulation of photoelectric signals and stores the signals of each defense zone as one-dimensional timing signals. The signal analysis module deploys the proposed method by our research on the upper computer for signal processing and pattern recognition. After providing early warning information, UAVs are dispatched for verification and intervention (\textcircled{5}).

When applying DOFS in the pipeline safety field, researchers initially focused on detecting external intrusions to the pipeline \cite{r40, ryangaaai}. Due to the fact that the DAS system continuously monitors the pipeline, with the sensing fibers often laid parallel to the pipeline, it is relatively easy to locate the occurrence position of intrusion events along the axial direction of the pipeline \cite{r41}. Tejedor et al. made a series of works by combining DOFS with machine learning methods in the field of pipeline safety warnings. Furthermore, researchers have increasingly paid attention to the types of external events in the axial direction of the pipeline and conducted research in the direction of pattern recognition \cite{r42,r20}. Pipeline axial localization can estimate the occurrence of external damage events of pipelines, so as to give early warning to threatening events. However, in addition to the model's performance issues, the radial distance of external hazardous events (such as mechanical construction) from the pipeline also caused a large number of false positives.

In the past, few researches conducted on the distance of hazardous events occurring in the radial direction of pipelines \cite{r3}. Nevertheless, optical fibers have a wide range of external sensing, which can generally sense vibration signals along the radial direction of the pipeline at around 50 meters. According to relevant industry standards and regulations, the safety distance for construction near oil and gas pipelines is typically 5 meters, while the safety distance for structures such as factories is 22.5 meters. Moreover, we classify the areas with different radial distances into three classes as shown in Fig. \ref{fig_1}. No processing is required for the non-threat area. For the tracking area, we can continue monitoring the threatening behavior, and further determine its working condition in the next time step. For the alarm area, we need to take timely verification and intervention measures, such as sending UAVs for evidence.

\section{Methodology}

\subsection{Signal Preprocessing and Feature Construction}

The signals collected by the DAS system are of low signal-to-noise ratio due to both device-generated noises and environmental noise, resulting in unsatisfactory signal quality. To address this issue, we first perform a pre-processing operation of denoising and reconstructing the signals. Specifically, we use variational mode decomposition (VMD) to enhance the signal quality, thereby improving the performance of the feature extraction and recognition network. VMD employs a non-repeated development process, avoiding modal distortion and ensuring operational efficiency \cite{rvmd,rpyy}.

Then, we mainly introduce the feature construction method proposed in this paper. Our approach is built on the basis of a signal time-frequency map and integrates multi-view and multi-domain features from both the time-domain, frequency-domain and space-domain of continuous defense zones. As shown in Fig. \ref{fig_2}, the vibration generated by external events has an effect on the axial localization defense zone and its adjacent defense zones. The location information of the radial direction can be hidden within the acoustics signals of adjacent defense zones to the axial localization defense zone. Therefore, we plot time-frequency maps \cite{rtf1} for each $i^{th}$ defense zone and its adjacent zone $i-1$ and $i+1$. Finally, we obtain a space-time-frequency fusion feature map through image stitching.

We perform this approach to achieve the same as multi-point spatial localization. We found that limited dimensional feature extraction results in significant feature overlap when attempting to achieve fine-grained levels of recognition tasks (such as radial localization). It is difficult to distinguish radial location solely through the signal from a single defense zone, while the complementary information from adjacent defense zones can enhance behavioral features in space. The vibration transmission of excavating behavior and the fading law of sensing light can enable nearby defense zones to capture effective information on threat signals \cite{rmulit, raaai2}. Additionally, on the advice of energy pipeline experts, two adjacent defense zones are selected and the feature fusion of three zones proved to be a satisfactory result. Increasing the adjacent defense zone signals would result in redundant information, increased computational complexity, and reduce processing speed. Therefore, we construct features for threat signals as shown in Fig. \ref{fig_2} to better distinguish signals from three radial threat areas.

\begin{figure}[!t]
\centering
\includegraphics[width=1.0\columnwidth]{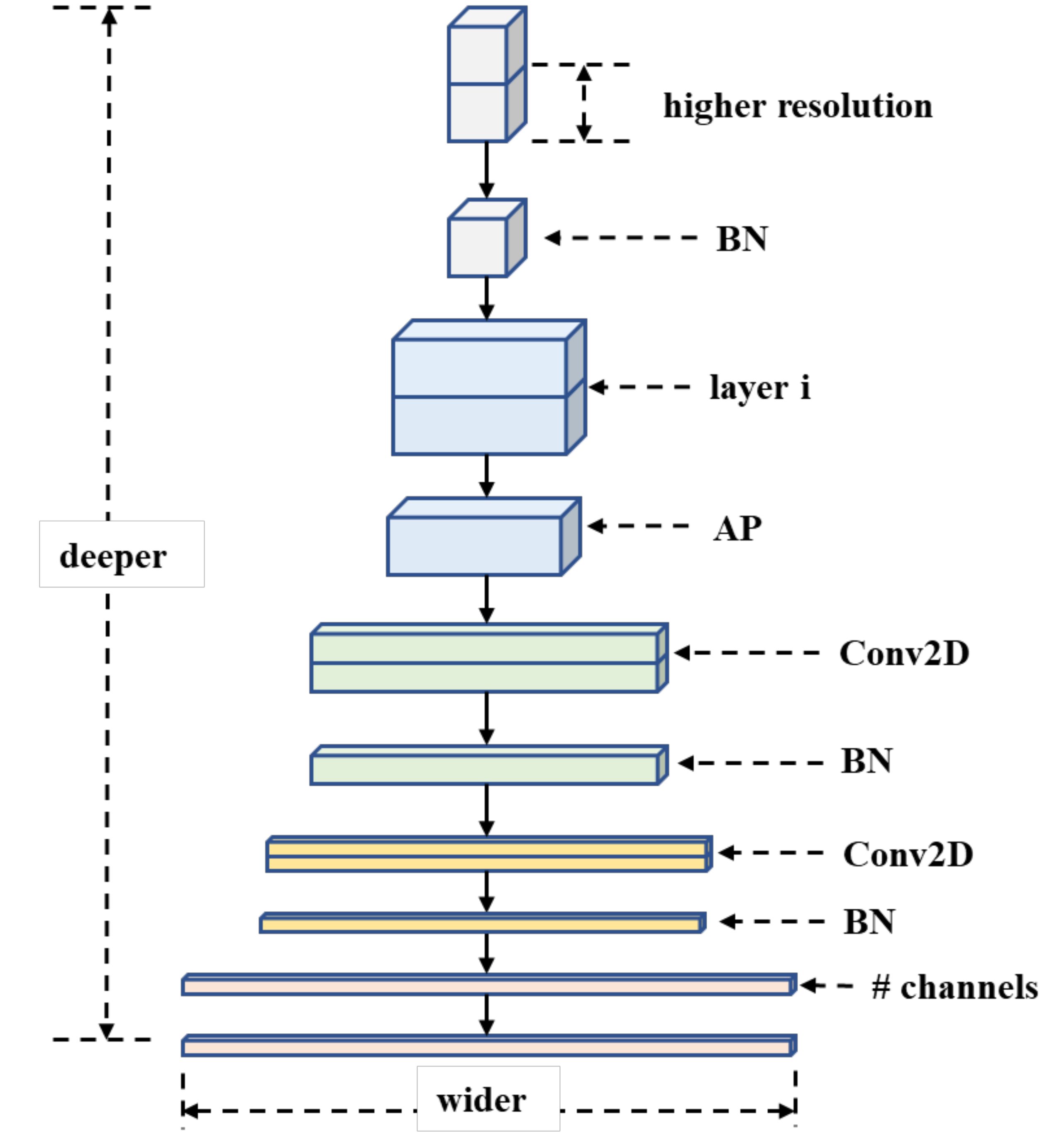}
\caption{A composite scaling method that uniformly scales all three dimensions at a fixed ratio.}
\label{fig_3}
\end{figure}

\begin{figure*}[!t]
\centering
\includegraphics[width=0.943\textwidth]{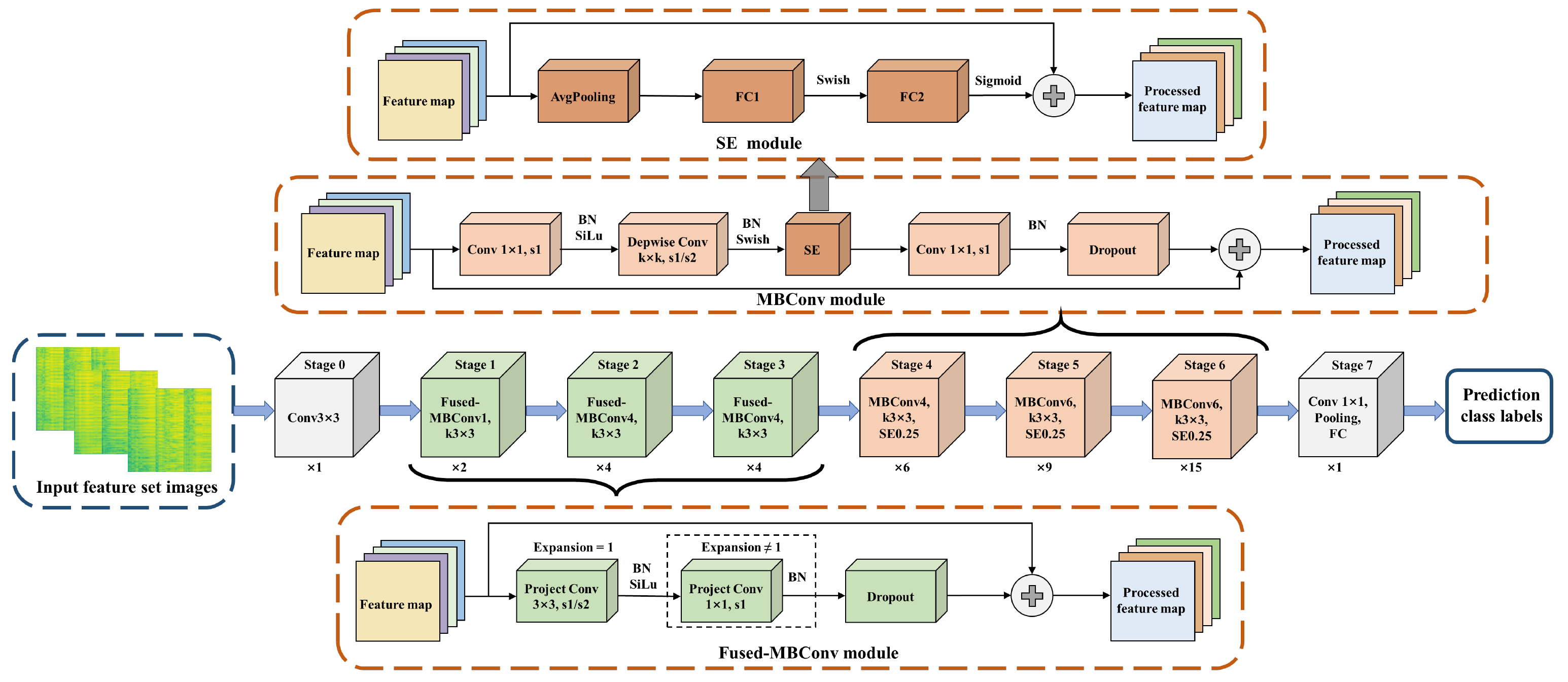}
\caption{The structure of pipeline radial threat estimation and recognition network.}
\label{fig_4}
\end{figure*}

\subsection{Threat Estimation and Recognition Network}

Considering that radial threat estimation is performed on the basis of axial localization, from the perspective of the practical application of AI technology, we need to balance model performance and time consumption when building the recognition network. Therefore, lightweight design has more advantages. We choose to build the threat area recognition network based on EfficientNetV2 \cite{reffv2}.
EfficientNet comprehensively optimizes the effects of input resolution, network depth, and network width to obtain the optimal network parameters for specific tasks. Especially, a composite scaling method that uniformly scales all three dimensions at a fixed ratio (enhancing network width, depth, and resolution) is proposed. As shown in Fig. \ref{fig_3}, a convolutional network layer $i$ can be defined as: 

\begin{equation}
\label{deqn_ex1}
Y_{i}=F_{i}\left(X_{i}\right)
\end{equation}

\noindent where $F$ denotes the operation, $Y$ represents the output tensor, $X$ is the input tensor with dimension $<H_{i}, W_{i}, C_{i}>$. $H$ and $W$ are the spatial dimension, and $C$ is the channel dimension. $H_{i}$, $W_{i}$, and $C_{i}$ are predetermined parameters in the baseline network. CNN can define the same or value of each convolution calculation in the network as:

\begin{equation}
\label{deqn_ex2}
N=F_{k} \odot \ldots \odot F_{2} \odot F_{1}\left(X_{1}\right)=\odot_{i} F_{i}^{d L_{i}}\left(X_{\left(r H_{i}, r W_{i}, w C_{i}\right)}^{i}\right)
\end{equation}

\noindent where $F_{i}^{d L_{i}}$ indicates that $F_{i}$ is repeated $L_{i}$ times in the stage $i$. $<H_{i}, W_{i}, C_{i}>$ represents the type of input tensor in the $i^{th}$ layer, and $w$, $d$, and $r$ are coefficients for scaling network width, depth, and resolution. $F_{i}$ and $L_{i}$ are also predetermined parameters in the baseline network.

The overall network structure of EfficientnetV2 is shown in Fig. \ref{fig_4}. It mainly consists of MBconv and Fused-MBConv modules. In Fig. \ref{fig_4}, we specifically demonstrate the operation process of these two modules. The MBconv module includes an attention mechanism of sequence and extraction (SE) block \cite{rse}. The purpose of the SE module is to obtain more important feature information through a weight matrix. Eq. 3 to 5 are performed on the feature map to obtain different weights assigned to different positions of the image from the channel domain.

\begin{equation}
\label{deqn_ex3}
z_{c}=\mathbf{F}_{s q}\left(\mathbf{u}_{c}\right)=\frac{1}{H \times W} \sum_{i=1}^{H} \sum_{j=1}^{W} u_{c}(i, j)
\end{equation}

\begin{equation}
\label{deqn_ex4}
\mathbf{s}=\mathbf{F}_{e x}(\mathbf{z}, \mathbf{W})=\sigma(g(\mathbf{z}, \mathbf{W}))=\sigma\left(\mathbf{W}_{2} \delta\left(\mathbf{W}_{1} \mathbf{z}\right)\right)
\end{equation}

\begin{equation}
\label{deqn_ex5}
\widetilde{\mathbf{X}}_{c}=\mathbf{F}_{s c a l e}\left(\mathbf{u}_{c}, s_{c}\right)=s_{c} \mathbf{u}_{c}
\end{equation}

\noindent where $\mathbf{F}_{s q}(\cdot)$ represents global average pooling. $H$, $W$ and $C$ represent the height, width and channel respectively. $\mathbf{F}_{e x}(\cdot)$ represents two fully connected operations. $\sigma$ and $\delta$ are the activation functions Sigmoid and ReLU respectively. $W$, $W_{1}$ and $W_{1}$ represent the weight. Both $\widetilde{\mathbf{X}}_{c}$ and $\mathbf{F}_{s c a l e}(\cdot)$ mean the channel multiplication between the metric quantity $s_{c}$ and the feature map $u_{c}$.

For the Fused-MBConv module, when expansion $=$ 1, there is only one 3$\times$3 convolution on the main branch, followed by the BN layer and SILU activation function, and a Dropout. Moreover, when expansion $\neq$ 1, there is an ascending dimension of 3$\times$3 convolution on the main branch, followed by the BN layer and SILU activation function, and then pass through 1$\times$1 convolution, BN and Dropout.

\subsection{Pre-training Model Based on Transfer Learning}

Transfer learning refers to the ability to transfer knowledge from one domain to another. During training a neural network, it acquires knowledge information and converts it into corresponding weights \cite{rtl}. These weights can be extracted and transferred to a new network model in the emerging field. In computer vision, transfer learning is typically achieved by using a pre-trained model. It is a model trained on large benchmark datasets to solve similar problems. The research on pipeline radial threat estimation faces challenges such as difficulty in obtaining data samples and limited dataset size. Therefore, transfer learning based on pre-trained models can improve the learning efficiency and performance of the model.

\begin{figure}[!t]
\centering
\includegraphics[width=1.0\columnwidth]{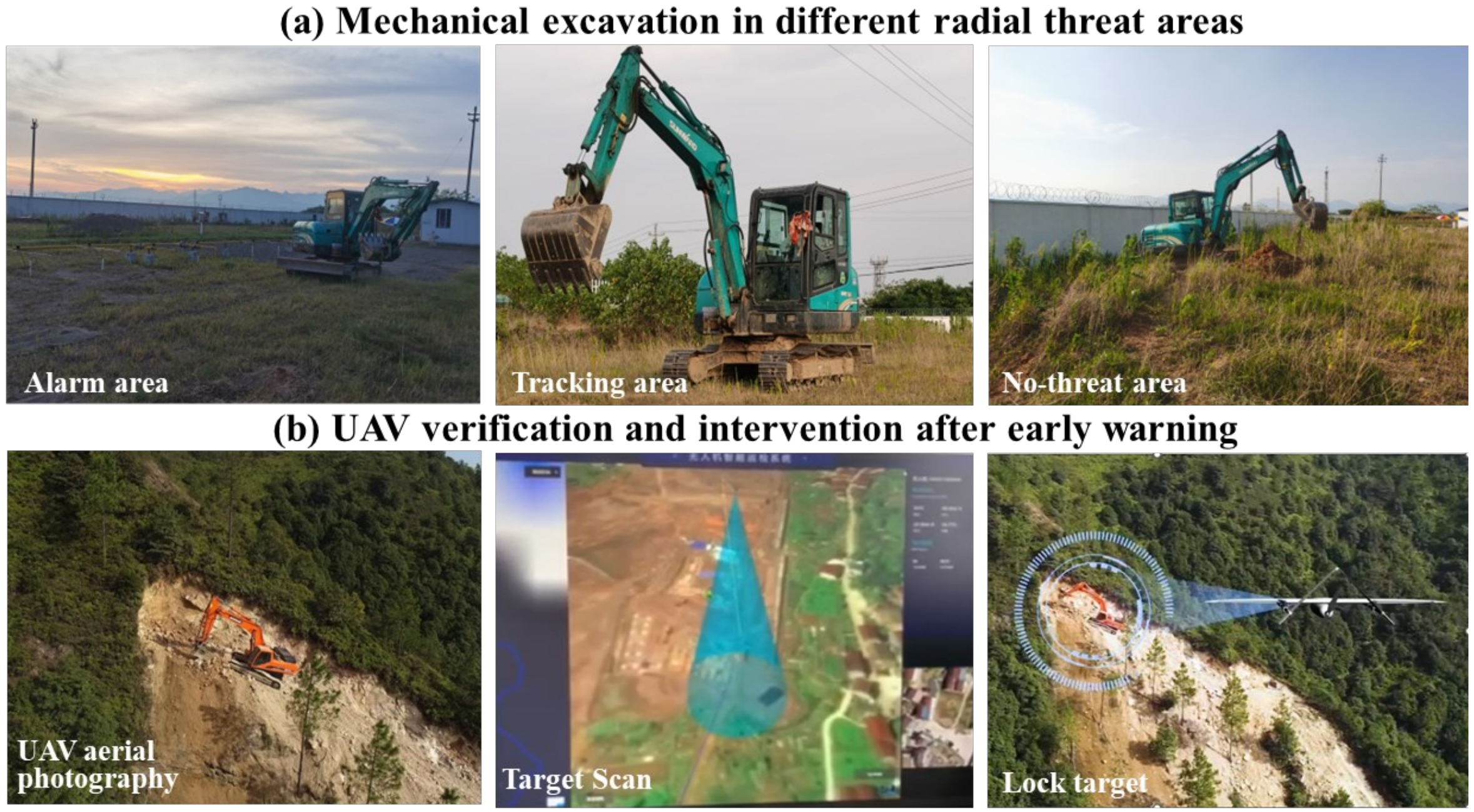}
\caption{Experimental data collection in the real-world scenario. (a) Mechanical excavation in different radial threat areas. (b) UAV verification and intervention after early warning.}
\label{fig_5}
\end{figure}

\section{Experiments and Analysis}

The section introduces the dataset used to verify the effectiveness of our work, as well as the comparative experiments that we set up, evaluation indicators, implementation details, and the analysis and discussion of the experimental results.

\subsection{Data Acquisition}

\begin{figure*}[!t]
\centering
\includegraphics[width=0.943\textwidth]{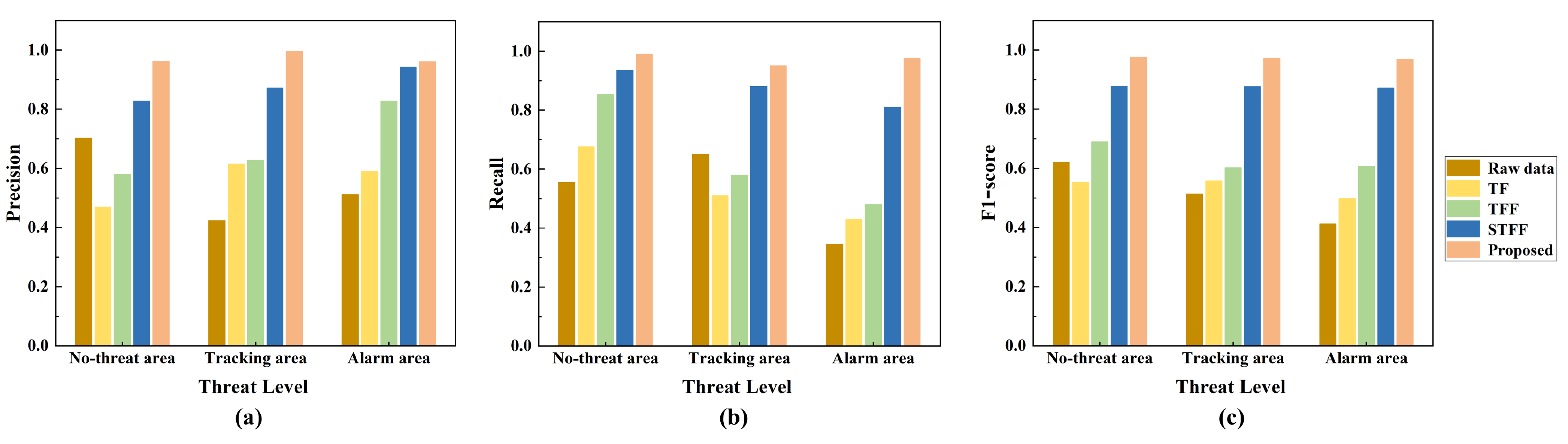}
\caption{The comparison of per-class evaluation indicators for the five processing methods. (a) precision. (b) recall. (c) F1-score.}
\label{fig_6}
\end{figure*}

Due to the uniqueness of the research domain and the novelty of the research issue, there are no relevant publicly available datasets that can be utilized. Therefore, we conduct on-site experiments in real-world scenarios to collect experimental data for validating our proposed method. As shown in Fig. \ref{fig_5} (a), we carry out signal acquisition for mechanical excavation events at different radial distances (5m, 20m, and 40m) in a certain deployed natural gas pipeline network in Zhejiang Province, China. The setting of three threat areas categories (alarm area, tracking area, and non-threat area) is based on the requirements for construction safety distances for pipelines as per relevant global (industry) standards \cite{rs1}. Fig. \ref{fig_5} (b) shows that we dispatch UAVs for verification and intervention for the true early warning from an intelligent sensing system.

The device used is a self-developed DAS system. Some basic parameters of the system include the monitoring distance of 20 km (bidirectional), the spatial resolution set to 10 m, and the sampling frequency of 2 kHz. By dispatching excavator machines to simulate threat scenarios, the signals collected can truly reflect the characteristics of the real site. It is evident that such an experiment is expensive, including the suspension of natural gas transmission and operation, as well as costs related to machinery and labor. Therefore, our dataset is limited in size. After the screening, we obtained 300 effective samples for each of the three areas, covering complete data for every 2,000 defense zones (sensing range of 20 kilometers) within the experimental area. The selected effective data samples only retain the time-sequence signals from the defense zone of excavating operation and adjacent defense areas to establish the data set for the experiment. 

To ensure the performance of the model, we augment the dataset by increasing the sample set size. Meanwhile, to avoid the influence of data imbalance on the model, each type of sample has a quantity of 1,000 \cite{rda}. This is a feature dataset that integrates expert knowledge\footnote{https://github.com/zhuchengyuan517/AAAI-7856-DATA.git}. Each sample data consists of 20,000 points, representing the sensing signal of the $i^{th}$ construction area in 10 seconds. The labels come from the labeling (threat distance) when conducting experiments in the experimental areas. In addition, we randomly divide the dataset into training and testing sets according to a 4:1 ratio. 

\subsection{Experiment Setting}

To demonstrate the effectiveness of the proposed feature construction method, we conduct comparative experiments to compare the proposed method with four other processing methods, including raw data without any processing, time-domain feature (TF) extraction, time-frequency domain feature (TFF) extraction, and space-time-frequency domain feature (STFF) extraction without data enhancement. To ensure the validity of experimental results, the classifiers are all based on the EfficientNetV2 model \cite{reffv2}, and the model parameters are set according to the introduction in the previous sections. The model uses an SGD optimizer (set momentum = 0.9, weight decay = 1e-4). The batch size is 8 and 100 epochs are performed. Besides, the learning rate is 0.01 and the final learning rate is 0.001. In addition, we conduct model-level comparative experiments to demonstrate the advantages of the constructed recognition network and pre-trained model. Due to the consideration of training and recognition computational costs, the comparison network we established has the same level of depth, including MobileNetV2 \cite{rmobilev2}, EfficientNetV1 \cite{reffv1}, and Swin Transformer-B \cite{rswint}. Moreover, the performance of the model and the required training epoch are experimentally achieved without the use of pre-trained models. The signal preprocessing methods and recognition model are shown in the appendix\footnote{https://github.com/zhuchengyuan517/AAAI-7856-CODE.git}.

To demonstrate the superiority of our proposed method, we use per-class evaluation indexes (Precision, Recall and F1-score) and average evaluation indexes ($P_{ave}$, $R_{ave}$ and $F1_{ave}$) to comprehensively evaluate the model. Since the sample size for each class is the same, the values of accuracy and average recall are the same. Therefore, we only use Recall to compare different models between the two indexes. Additionally, we also select false alarm rate (FAR) as an equally important evaluation metric.

\subsection{Result and Analysis}

\begin{figure}[!t]
\centering
\includegraphics[width=1.0\columnwidth]{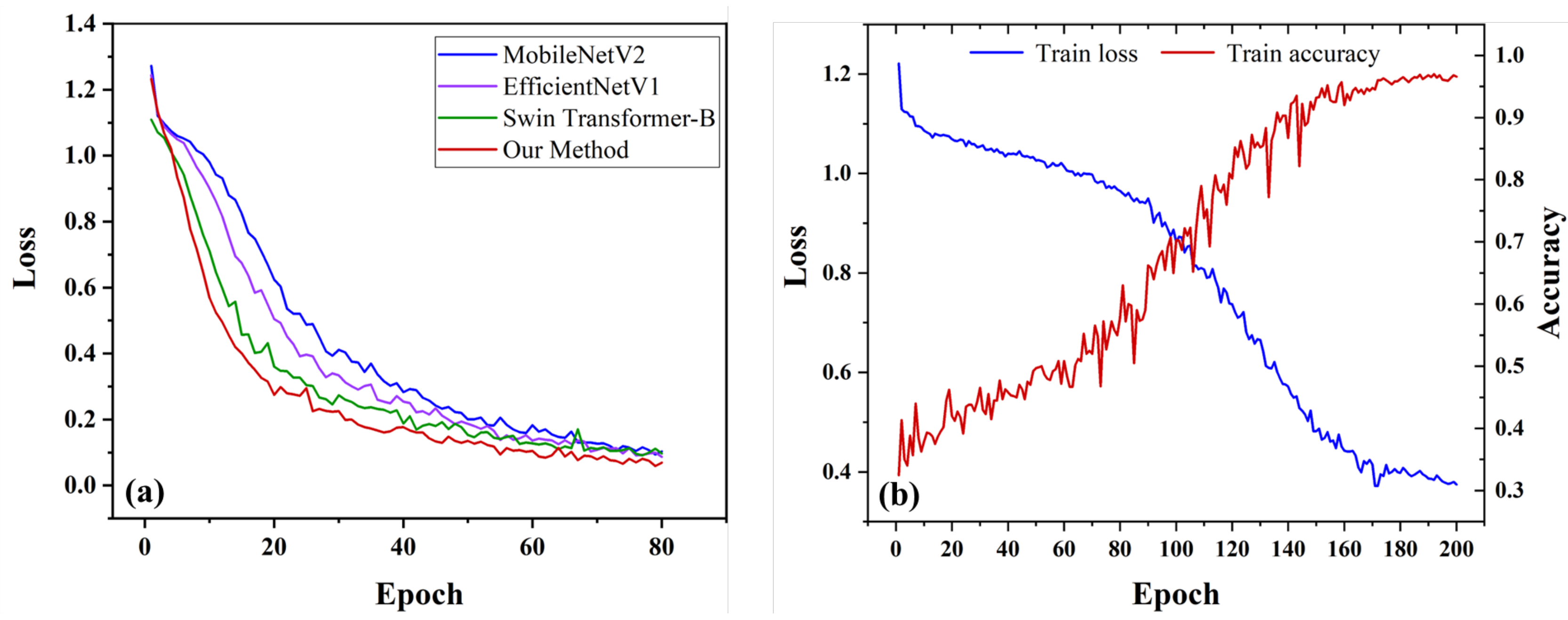}
\caption{The training curves for different models. (a) The training loss curve of four recognition models. (b) The training loss curve and accuracy curve of our model without the pre-trained model.}
\label{fig_7}
\end{figure}

The experimental results demonstrate that the feature set obtained from the multi-view and multi-domain fusion of continuous defense zones features significantly improves the model recognition performance, confirming our initial hypothesis. As shown in Fig. \ref{fig_6}, it represents the recognition results of different methods. It can be seen that the differentiation effect becomes significant when the multi-view and multi-domain features of continuous defense zones are fused. This is because the multi-view and multi-domain features can capture more detailed information about the samples, which helps to improve the model’s ability to distinguish between different types. 

\begin{table}[!t]
\centering
\resizebox{0.42\textwidth}{!}{
\begin{tabular}{ccccc}
\hline \hline
Method       & $P_{ave}$ $\uparrow$ (\%)             & $R_{ave}$ $\uparrow$  (\%)          & $F1_{ave}$ $\uparrow$  (\%)           & FAR  $\downarrow$  (\%)         \\ \hline
Raw data     & 54.57          & 51.67          & 51.50          & 65.92          \\
TF           & 55.74          & 53.83          & 53.59          & 44.52          \\
TFF          & 67.80          & 63.75          & 63.33          & 26.72          \\
STFF         & 88.02          & 87.50          & 87.48          & 5.81          \\
Ours & {\textbf{97.82}} & {\textbf{97.67}} & {\textbf{97.69}} & {\textbf{0.99}} \\ \hline \hline
\end{tabular}}
\caption{The comparison of comprehensive indexes of the five processing methods.}
\label{table1}
\end{table}

As shown in Table 1, the model performance is greatly improved with significant improvements in $P_{ave}$, $R_{ave}$, and $F1_{ave}$ for all three types using the STFF method. Furthermore, the proposed data augmentation techniques further enhance the model learning performance, resulting in improved recognition performance. It can be seen that the $P_{ave}$ of the proposed method reaches 97.82\%, with a significant advantage, 9.8\% higher than the STFF without data augmentation of the best performance of other methods. $R_{ave}$ and $F1_{ave}$ are also high, reaching more than 97\%, which indicates that the model has good generalization ability. Especially, compared with the other four methods, the FAR of the proposed method reaches the best effect of 0.99\%. The experimental results we presented are averaged over 10 repeated experiments. We found the misclassified samples are all divided into tracking areas. This is a good sign because the tracking area also represents the potential threat. Moreover, the reason for this phenomenon may be that  some samples in the alarm area are very close to the tracking area.

\begin{table}[!ht]
\resizebox{1.\columnwidth}{!}{
\centering
\begin{tabular}{ccccc}
\hline \hline
\multirow{2}{*}{Model} & \multicolumn{3}{c}{Class}                           & \multirow{2}{*}{\begin{tabular}[c]{@{}c@{}}Comprehensive \\ index\end{tabular}} \\ \cline{2-4}
                       & No-threat area  & Tracking area   & Alarm area      &                                      \\ \hline
                       & \multicolumn{3}{c}{Precision $\uparrow$ (\%) }                       & $P_{ave}$  $\uparrow$ (\%)                                    \\ \hline
MobileNetV2            & 89.22          & 91.05          & 90.78          & 90.35                                                                          \\
EfficientNetV1         & 97.95          & {\textbf{100}} & 88.39          & 95.45                                                                          \\
Swin Transformer-B     & 93.63          & 97.31          & 92.38          & 94.44                                                                          \\
Our method             & {\textbf{100}} & 93.46          & {\textbf{100}} & {\textbf{97.82}}                                                                 \\ \hline
                       & \multicolumn{3}{c}{Recall $\uparrow$ (\%) }                          & $R_{ave}$   $\uparrow$ (\%)                                                                             \\ \hline
MobileNetV2            & 91.00          & 86.50          & 93.50          & 90.33                                                                          \\
EfficientNetV1         & 95.50          & 90.50          & {\textbf{99.00}} & 95.00                                                                          \\
Swin Transformer-B     & 95.50          & 90.50          & 97.00          & 94.33                                                                          \\
Our method             & {\textbf{96.00}} & {\textbf{100}} & 97.00          & {\textbf{97.67}}                                                                 \\ \hline
                       & \multicolumn{3}{c}{F1-score $\uparrow$ (\%) }                        & $F1_{ave}$  $\uparrow$ (\%)                                                                             \\ \hline
MobileNetV2            & 90.01          & 88.72          & 92.12          & 90.31                                                                          \\
EfficientNetV1         & 96.71          & 95.01          & 93.40          & 95.04                                                                          \\
Swin Transformer-B     & 94.55          & 93.78          & 94.63          & 94.32                                                                          \\
Our method             & {\textbf{97.96}} & {\textbf{96.62}} & {\textbf{98.48}} & {\textbf{97.69}}                                                                 \\ \hline \hline
\end{tabular}
}
\caption{The comparison of evaluation indicators for four network models.}
\label{table2}
\end{table}

Fig. \ref{fig_7} (a) reports the training process of the network loss functions for four recognition models. It can be seen that our proposed model can converge faster. Additionally, we compare the training of the model without the pre-trained model, as shown in Fig. \ref{fig_7} (b). It takes 200 epochs to train the model before reaching the effect achieved by only 40 epochs using a pre-trained model. Table 2 shows the per-class and comprehensive evaluation indicators of different recognition models. It can also be seen from the table that our model has advantages in various indicators. Among them, $P_{ave}$, $R_{ave}$ and $F1_{ave}$ all reach over 97\%, higher than the other three models. Furthermore, we report on the params, FAR and the testing time consumption of the comparison's four models, as shown in Table 3. It can be seen that our method achieves the lowest FAR and the shortest response time. In summary, our proposed method can provide more effective information for energy pipeline operators to identify true threat events in practical scenarios. Our system based on the above-proposed algorithm deployment on over 2000 kilometers of pipelines can reduce false alarms by more than 10,000 annually, saving tens of millions of costs in operation and maintenance management.

\begin{table}[!t]
\centering
\resizebox{.9\columnwidth}{!}{
\begin{tabular}{cccc}
\hline \hline
Model              & Params  $\downarrow$                                                            & FAR $\downarrow$              & Time consumed $\downarrow$    \\ \hline
MobileNetv2        & {\textbf{17M}} & 9.22\%          & 3.5486s          \\
EfficientNetV1     & 30M       & 2.05\%          & 3.3737s          \\
Swin Transformer-B & 88M                                                                & 7.62\%          & 3.6328s          \\
Our Method         & 22M    & {\textbf{0.99\%}} & {\textbf{2.7245s}} \\ \hline \hline
\end{tabular}
}
\caption{The network parameters and performance indicators of four models.}
\label{table3}
\end{table}

\section{Conclusion}
In this paper, we propose a radial threat estimation and early warning method based on distributed optical fiber sensing systems, which are used to monitor and identify real external threat events of energy pipelines. In detail, we design a novel feature construction method that integrates multi-view and multi-domain features from continuous defense zones. Moreover, the composite feature extraction strategy and recognition model construction have been demonstrated to be more advantageous. This approach deepens the sensing information and increases the recognition dimension, thereby improving the effectiveness of safety decision-making for pipelines. Through the validation of real-world datasets, our proposed method achieves state-of-the-art performance while greatly reducing false alarm rates. Furthermore, our research not only meets the practical needs of various energy scenarios but also allows for further exploration of fine-grained granularity in the future, such as signal tracking for disaster prevention and reduction, condition identification in industrial scenarios, and perimeter intrusion behavior monitoring. The proposed pipeline safety early warning algorithm and system are already in use in real sites and have a positive impact on energy security and societal stability.

\section{Acknowledgments}

This work was supported in part by the "Pioneer" and "Leading Goose" R\&D Program of Zhejiang under Grant 2022C01178, in part by the National Natural Science Foundation of China No. U21A20478, No. 62106224, and No.62002322, in part by Zhejiang High-Level Talents Special Support Program under Grant 2021R52002, in part by Zhejiang Provincial Nature Science Foundation of China under Grant LZ21F030004, and in part by the Key Research and Development Program of Zhejiang Province under Grant 2021C03037.

\section{Ethical Statement}
This paper proposes a novel external safety warning method for energy pipeline transportation and applies it in a real-world scenario. In terms of ethical and social implications, our method incorporates artificial intelligence and intelligent sensing systems to help the energy industry monitor the safety of energy pipelines with a more fine-grained perspective, significantly improving the intelligence of the sensing system in this field. Therefore, it can not only ensure the transmission and supply of energy but also reduce the consumption of resources for the purpose. On the other hand, due to the current stage of artificial intelligence development, we are constantly exploring its enormous benefits to society. However, it may also present unexpected situations. Although our proposed system has been verified in practical applications, it still has some uncertainties such as misoperation, insufficient power supply or communication system failures. Considering the importance of the energy industry, the best solution is to arrange for a few professionals and experts to supervise and optimize the system, thereby guaranteeing the validity of our solution and maximising the system's benefit to society.

\bibliography{aaai24}

\end{document}